\documentclass[letterpaper]{article}
\usepackage{aaai2026}
\usepackage{times}
\usepackage{helvet}
\usepackage{courier}
\usepackage[hyphens]{url}
\usepackage{graphicx}
\usepackage{natbib}
\usepackage{caption}
\usepackage{algorithm}
\usepackage{algorithmic}

\usepackage{amsmath}
\usepackage{subcaption}
\usepackage{amsfonts}

\usepackage{newfloat}
\usepackage{listings}
\DeclareCaptionStyle{ruled}{labelfont=normalfont,labelsep=colon,strut=off} 
\floatstyle{ruled}
\newfloat{listing}{tb}{lst}{}
\floatname{listing}{Listing}
\title{
Probing the Origins of Reasoning Performance: Representational Quality for Mathematical Problem-Solving in RL vs.\ SFT Fine-Tuned Models
}

\author {
    \begin{tabular}{c@{\hspace{2.5em}}c@{\hspace{2.5em}}c}
        \rule{0pt}{2.6em}Antyabha Rahman & Akshaj Gurugubelli & Omar Ankit \\
        {\normalsize\normalfont University of New South Wales} & {\normalsize\normalfont Algoverse AI Research} & {\normalsize\normalfont University of Waterloo}
    \end{tabular}\\[1.2ex]
    \begin{tabular}{c@{\hspace{2.5em}}c}
        Kevin Zhu & Aishwarya Balwani \\
        {\normalsize\normalfont Algoverse AI Research} & {\normalsize\normalfont St. Jude Children's Research Hospital} \\[1.5em]
    \end{tabular}
}
\affiliations {
}

\begin{document}
\maketitle

\begin{abstract}
    Large language models trained with reinforcement learning (RL) to reason step by step have been increasingly shown to outperform their supervised fine-tuned (SFT) counterparts on mathematical reasoning tasks.
    Yet the mechanistic basis for this improvement remains unclear.
    We therefore ask, \textit{what internal representational differences enable RL-trained reasoning language models to reason better?}
    Our work presents two primary lines of evidence:
    First, layer-wise linear probes tend to predict answer correctness more accurately in RL-trained reasoning language models than in instruction-tuned models on synthetic math problems, although output length alone matches the DeepSeek-Math-RL probe on two templates.
    Second, mean ablation studies suggest that accuracy drop rises with layer depth in DeepSeek-Math-RL but not in DeepSeek-Math-Instruct.
    Finally, we analyse token-count variability across accuracy bins for RL-trained reasoning language models and instruction-tuned models to measure whether token counts vary more on harder problems. Token counts tend to vary more on lower-accuracy problems, although in DeepSeek-Math-Instruct only with truncated responses included.
    At high accuracy, we observe lower variability in DeepSeek-Math-RL than in DeepSeek-Math-Instruct, and we observe higher variability in Olmo-3-Think than in Olmo-3-Instruct.
\end{abstract}
    
\begin{links}
    \link{Code}{https://oankit.github.io/-rl-sft-reasoning/}
\end{links}

\noindent\textbf{Correction note.} This version corrects arXiv:2607.26119v1. The Correction Note before the appendices lists the main changes.

\section{Introduction}
    
Language models trained with reinforcement learning (RL) to reason step by step, such as OpenAI's o1~\cite{openai2024openaio1card} and DeepSeek-R1, outperform supervised fine-tuned (SFT) language models on verifiable reasoning tasks such as mathematics and coding~\cite{deepseekai2025deepseekr1incentivizingreasoningcapability}.
Understanding \textit{why} this is the case requires moving beyond performance metrics to mechanistic explanations. While we know RL-trained reasoning language models achieve higher accuracy, \textit{how} they differ internally from SFT language models remains an open challenge.

Current research has approached this question from two mutually reinforcing but disconnected perspectives. \textbf{Mechanistic interpretability} has identified specific circuits for arithmetic operations~\cite{hanna2023doesgpt2computegreaterthan}, used probes to read arithmetic errors and number values~\cite{sachan2025probingarithmeticerrors,zhu2024languagemodelsencodevalue}, and shown that chain-of-thought increases activation sparsity~\cite{chen2025doeschainthoughtthink}, primarily in smaller models on elementary operations. \textbf{Behavioural studies} have revealed information-theoretic compression limits~\cite{lee2025llmscompresschainofthoughttoken}, cross-variant sensitivity to problem phrasing~\cite{mirzadeh2025gsmsymbolicunderstandinglimitationsmathematical}, and RL training's potential for long reasoning~\cite{yeo2025demystifyinglongchainofthoughtreasoning}.
However, \textit{what internal representational differences} enable RL-trained reasoning language models' problem-solving performance remains unexplored.

\textbf{We seek to bridge this gap} through integrated behavioural-mechanistic analysis using three complementary methods:
(1)~\textit{Linear probing} on layer-wise hidden states to predict answer correctness~\cite{alain2018understandingintermediatelayersusing,belinkov-2022-probing}, measuring \textit{when} and \textit{how strongly} representations of correctness emerge across layers and model families.
(2)~\textit{Mean ablation interventions}~\cite{zhang2024bestpracticesactivationpatching,meng2023locatingeditingfactualassociations} to identify which layers are critical for mathematical reasoning across training methodologies.
(3)~\textit{Token-count variability analysis} via multiple responses per problem with temperature and top-p matched across models, extending variance analysis~\cite{mirzadeh2025gsmsymbolicunderstandinglimitationsmathematical} to within-problem comparisons and empirically testing compression theory~\cite{lee2025llmscompresschainofthoughttoken}.

Our contributions include:
(1) Evidence from scalable linear probing that RL-trained reasoning language models tend to score higher in probe accuracy than instruction-tuned models, with output length and label effects as confounds.
(2) Preliminary evidence that accuracy drop under mean ablation rises with layer depth in DeepSeek-Math-RL but not in DeepSeek-Math-Instruct.
(3) An empirical test of compression theory, measuring token-count variability across accuracy bins.
In DeepSeek-Math-RL, accuracy drop under mean ablation rises with layer depth (r=0.47), whereas DeepSeek-Math-Instruct shows no such trend (r=-0.11).
This difference, combined with higher probe accuracy in RL-trained reasoning language models on some problem templates, may provide initial mechanistic insight into performance differences, moving from descriptive benchmarking
\cite{openai2024openaio1card,deepseekai2025deepseekr1incentivizingreasoningcapability} toward mechanistic explanation.

\section{Measuring Representation Quality via Probing}
Understanding \textit{when} and \textit{how} correct answer information emerges across model layers can reveal fundamental differences between RL-trained reasoning language models and instruction-tuned models.
We evaluate four models spanning two architectural families, the DeepSeek-Math family~\cite{deepseek-math} and the Olmo-3 family~\cite{olmo2025olmo3}. The RL-trained reasoning language models, DeepSeek-Math-RL (DeepSeek-Math-Instruct further trained with GRPO and a learned reward model) and Olmo-3-Think (RLVR), have reasoning-specific RL training. Their instruction-tuned counterparts are DeepSeek-Math-Instruct and Olmo-3-Instruct. Both Olmo-3 models go through SFT, DPO, and RLVR, so only the DeepSeek pair contrasts RL with SFT.
We hypothesise that the improved reasoning ability of RL-trained reasoning language models on mathematical benchmarks stems from developing \emph{clearer} internal representations of correctness, i.e., representations that are more linearly separable and consistently structured across examples.
If these models develop ``clearer'' representations, we should be able to detect this mechanistically. Problems with more linearly separable internal representations should exhibit higher discriminability between correct and incorrect answers.
To test this, we train linear probes on layer-wise hidden states to predict final answer correctness, using probe accuracy as a proxy for representation clarity. We build on the interpretability literature using linear probes to study intermediate representations~\cite{alain2018understandingintermediatelayersusing}, particularly for mathematical reasoning~\cite{zhu2024languagemodelsencodevalue}.
We ask whether probe accuracy correlates with model accuracy, which could help explain why RL-trained reasoning language models outperform instruction-tuned models.

\subsection{Synthetic Problem Generation}
To investigate whether model failures stem from reasoning limitations or surface-form artefacts, we generate 1,000 synthetic mathematical problems for each of five fixed templates covering probability, fractions, and cost calculations. The templates are Cost Calculation, Sequential Growth, Counting, Student Demographics, and Conditional Probability. Template details and example problems are provided in Appendix~\ref{sec:linprobeexperiment}.
Following \citet{mirzadeh2025gsmsymbolicunderstandinglimitationsmathematical}'s approach of generating synthetic GSM8K variants~\cite{cobbe2021gsm8k}, our controlled generation isolates representational properties required for general reasoning from question-specific memorisation effects.

Synthetic generation offers three advantages. It (1) reduces data contamination, since the templates are numeric variants of GSM8K test problems, (2) enables algorithmic verification with known ground-truth parameters, and (3) scales to large sample sizes for robust statistical analysis. Prior work demonstrates that synthetic benchmarks effectively reveal model reasoning capabilities~\cite{mirzadeh2025gsmsymbolicunderstandinglimitationsmathematical} while avoiding artefacts of human-authored datasets.

\subsubsection{Completion Generation and Labelling}

For each model, we generate a single completion for every problem using sampling (temperature $T = 0.7$ for DeepSeek-Math and $T = 0.6$ for Olmo-3, \textit{top p} $= 0.95$). We extract final answers from within \texttt{\textbackslash boxed\{\}} delimiters or, for DeepSeek-Math-RL, which rarely boxes its answer, from the final ``The answer is'' line. Outputs with neither, all cut off at the token limit, are graded on their last number (1,433 of 19,000). We label a completion correct if the extracted answer matches the ground truth (allowing a tolerance of $\pm$1 for rounding), and incorrect otherwise. Completions with no extractable answer are omitted (1 of 19,000 on the reported templates), and answers that do not parse as a number count as incorrect.

\textbf{Balancing procedure.} To ensure fair comparison across models with different accuracy distributions, we sample equal numbers of correct and incorrect examples per model and template, all of the minority class and an equal random draw of the majority class, repeated over 5 seeds. We split the data 70/15/15 into train/validation/test sets, stratified by label. Balanced sets differ in size by model and template, and range from 142 to 996 examples across the runs we report. A run is evaluable only if each test class has at least 10 examples.

\subsubsection{Activation Extraction}

For completed outputs, activation extraction captures the model's internal state at the point where it has completed reasoning and is about to write its final answer. The final number may already appear earlier in the output.

\textbf{Probe position.} We feed the saved prompt and completion token IDs back through the model, locate the delimiter among the completion tokens, and extract hidden states at the \texttt{\{} of the last \texttt{\textbackslash boxed\{}. DeepSeek-Math-RL boxes only 2.0\% of its outputs, so when no \texttt{\textbackslash boxed\{} appears we use the last token of the final ``The answer is:''. Outputs cut off at the token limit with no \texttt{\textbackslash boxed\{} use their last token, so they are probed mid-reasoning, at a different token. We choose this position for two reasons: (1) for completed outputs, it comes before the final answer, ensuring the model has completed reasoning before answer articulation, and (2) the delimiters tokenise consistently across examples, unlike final answers which may span variable token lengths.

\textbf{Layer-wise representations.} We process completions one at a time, so no padding is needed. For each completion, we perform a single forward pass with \texttt{output\_hidden\_states=True} and extract activations from all $L$ transformer layers. We collect \texttt{hidden\_states[1:L+1]}, the outputs of transformer blocks 1 through $L$, omitting \texttt{hidden\_states[0]} (input embeddings). The last of these is post-normalised, so we exclude it when reporting peak accuracy. Figures and results number the layers from 0, so layer 0 is the output of block 1. The collected activations yield tensors of shape $[L \times N \times D]$, where $N$ is the number of completions and $D$ is hidden dimension (4096).

\subsubsection{Probe Training}

For each layer $\ell \in \{1, \ldots, L\}$, we train a logistic regression probe $f_\ell: \mathbb{R}^D \rightarrow \{0, 1\}$ to predict binary answer correctness from the $D$-dimensional hidden state $\mathbf{h}_\ell$. The probe is defined as
\begin{equation}
f_\ell(\mathbf{h}_\ell) = \sigma(\mathbf{w}_\ell^\top \mathbf{h}_\ell + b_\ell)
\end{equation}
where $\mathbf{w}_\ell \in \mathbb{R}^D$ is the weight vector, $b_\ell \in \mathbb{R}$ is the bias term, and $\sigma(x) = 1/(1 + e^{-x})$ is the sigmoid function. We standardise each feature and use 5-fold cross-validation to select regularisation strength $C$ from $\{0.01, 0.1, 1.0, 10.0, 100.0\}$, with \texttt{class\_weight='balanced'} to handle residual class imbalance.

We report test set accuracy on the balanced data, averaged over 5 seeds, as the primary metric. Our methodology assumes the linear representation hypothesis~\cite{zhu2024languagemodelsencodevalue}, that if correct answer information exists in a layer's representations, a simple linear classifier should reliably extract it. Higher probe accuracy indicates more robust linear separability between correct and incorrect answer representations.

\begin{figure}[t]
\centering
\includegraphics[width=\linewidth]{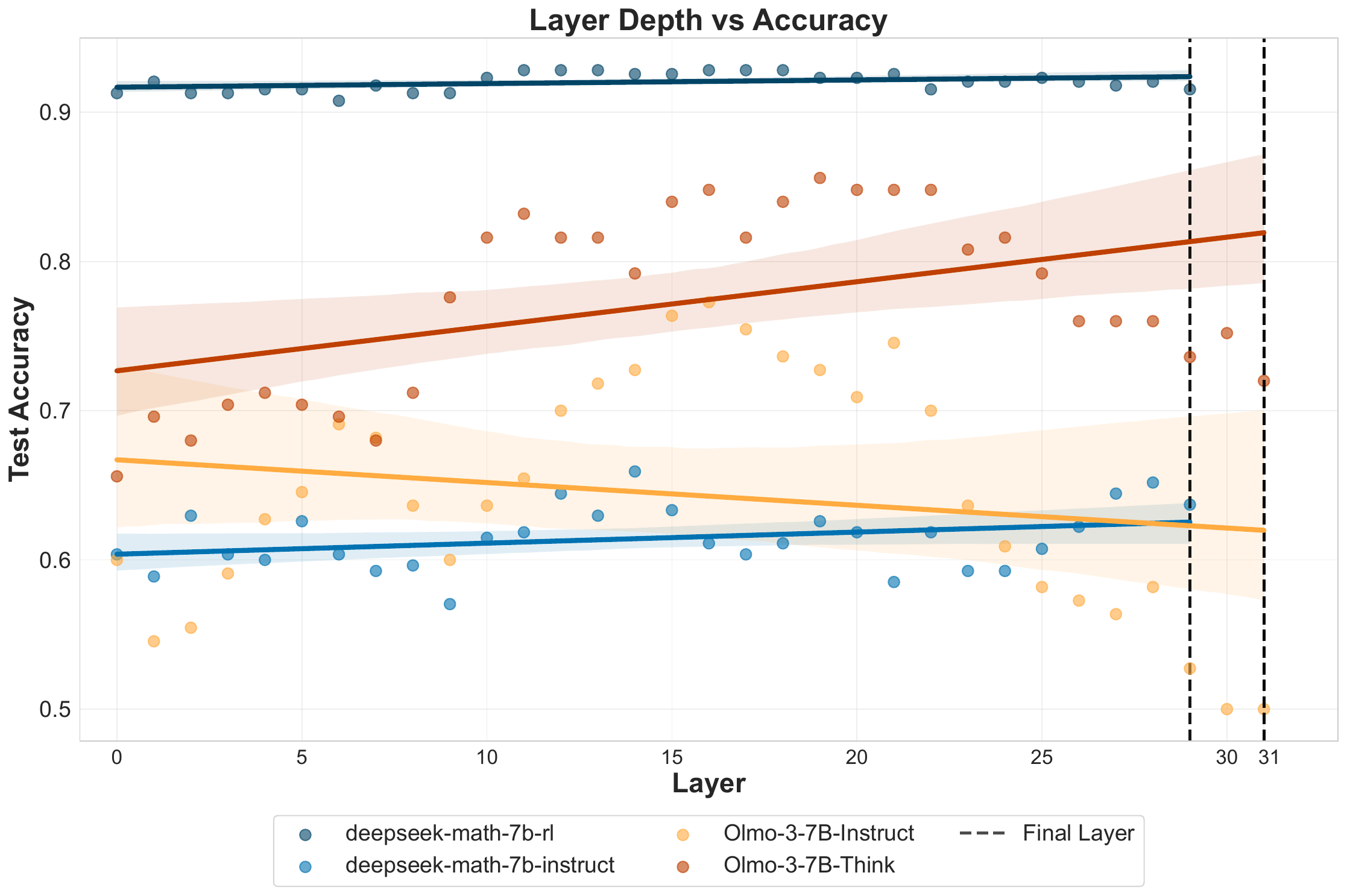}
\caption{Layer-wise probe accuracy for predicting answer correctness across model families on the Student Demographics template, on balanced data. Dashed lines mark each family's final, post-normalised layer. Late-layer regression appears for the Olmo-3 models. Output length and truncation alone give 93\% for DeepSeek-Math-RL on this template. Of the 82 Olmo-3-Think answers labelled incorrect, 80 are exactly 2 below a label that floors intermediate counts.}
\label{fig:layer_accuracy}
\end{figure}

\subsection{Results: Layer-Wise Probe Accuracy}

Figure~\ref{fig:layer_accuracy} shows differences in probe accuracy across model types. On Student Demographics, DeepSeek-Math-RL and Olmo-3-Think reach higher peak probe accuracy (93\% and 86\%) than their instruction-tuned counterparts (66\% and 77\%). Across the four templates with figures, peak accuracy is 65--98\% for DeepSeek-Math-RL and 65--92\% for DeepSeek-Math-Instruct. DeepSeek-Math-RL is higher on two of these templates, level with DeepSeek-Math-Instruct on Cost Calculation (65\% each), and lower on Conditional Probability (76\% against 81\%), although within the spread over seeds. Olmo-3-Think is higher than Olmo-3-Instruct on Cost Calculation (79\% against 56\%), and the two are within the spread over seeds on Student Demographics. A classifier on output length and truncation alone matches DeepSeek-Math-RL on Student Demographics, so these differences leave open whether RL training leads to representations with linear separability between correct and incorrect answer states~\cite{zhang2025reasoningmodelsknowtheyre,park2024linearrepresentationhypothesisgeometry}.

\textbf{Emergence across layers.} We define an \textbf{emergence layer} $\ell_{\text{emerge}}$ as the first layer achieving $>80\%$ test accuracy. On Student Demographics, $\ell_{\text{emerge}} = 0$ and $10$ for DeepSeek-Math-RL and Olmo-3-Think respectively, and DeepSeek-Math-Instruct and Olmo-3-Instruct do not reach this threshold. On this template, DeepSeek-Math-RL exhibits high probe accuracy from the very first layer, with test accuracy of 91\% at layer 0 compared to 66\% for Olmo-3-Think and 60\% for both Instruct models. DeepSeek-Math-RL and Olmo-3-Think average 92\% and 81\% over layers 15--29. Because output length alone matches the DeepSeek-Math-RL probe, its layer-0 accuracy does not show that a correctness representation emerges early.

\begin{table}[t]
\centering
\small
\setlength{\tabcolsep}{3pt}
\begin{tabular}{lr@{ / }lr@{ / }lr@{ / }lr@{ / }l}
\hline
Template & \multicolumn{2}{c}{DS-Inst.} & \multicolumn{2}{c}{DS-RL} & \multicolumn{2}{c}{Olmo-Inst.} & \multicolumn{2}{c}{Olmo-Think} \\
\hline
Cost & 52 & 65 & 59 & 65 & 88 & 56 & 91 & 79 \\
Growth & 83 & 92 & 84 & 98 & 95 & -- & 50 & 100 \\
Counting & 39 & 62 & 26 & 78 & 99 & -- & \multicolumn{2}{c}{--} \\
Demographics & 82 & 66 & 26 & 93 & 93 & 77 & 92 & 86 \\
Probability & 16 & 81 & 8 & 76 & 80 & 83 & 99.5 & -- \\
\hline
\end{tabular}
\caption{Model accuracy on all completions and peak probe accuracy on balanced data, in percent, as accuracy / probe. Accuracy uses the published $\pm$1 labels. A dash after the slash means too few examples to evaluate. Olmo-3-Think was not run on Counting.}
\label{tab:model_accuracy}
\end{table}

\textbf{Model accuracy.} Table~\ref{tab:model_accuracy} reports model accuracy next to peak probe accuracy. Probe accuracy does not track model accuracy (Spearman $\rho = 0.02$ across the 16 evaluable runs). RL-trained reasoning language models are also not always more accurate on these templates. DeepSeek-Math-RL is below DeepSeek-Math-Instruct on Counting and Student Demographics, where it truncates most outputs, and on Conditional Probability. Olmo-3-Think is below Olmo-3-Instruct on Sequential Growth, where it splits between two readings of the question, and level with it on Student Demographics. On GSM8K-Platinum the RL-trained reasoning language models are more accurate, 89\% against 82\% for DeepSeek-Math, and 98.5\% against 96.2\% for Olmo-3.

\textbf{Pre-training and training objectives.} Olmo-3-Instruct, post-trained from the base model on the Dolci Instruct data, outperforms DeepSeek-Math-Instruct. This suggests that a more robust and diverse dataset for pre-training and post-training can still provide significant performance increases. Our probe findings do not settle this, since the difference in probe accuracy for Olmo-3-Instruct changes sign across templates, and fine-tuning effects on probes can vary in size and direction~\cite{Mosbach2020OnTI}. DeepSeek-Math-RL and Olmo-3-Think score higher on probes than their instruction-tuned counterparts, beyond the spread over seeds, in 4 of 7 comparisons, but the confounds below mean this does not show that reasoning-specific training objectives, not merely scale or general fine-tuning, are key to developing clear correctness representations.

\subsection{Representation Clarity as a Possible Mechanism for Better Mathematical Reasoning}

Our results do not establish a mechanistic explanation for why RL-trained reasoning language models outperform instruction-tuned models on mathematical benchmarks. Clearer internal representations of correctness remain a hypothesis. The probe accuracy (56--100\%) varies by template and is affected by the limits below, so it does not show that these models encode correctness as robust, linearly-separable features~\cite{park2024linearrepresentationhypothesisgeometry,jiang2024originslinearrepresentationslarge}. If such clarity exists, it could enable more reliable access to correctness signals during autoregressive generation, leading to more consistent correct outputs.

\textbf{Caveats.} Our probing methodology has an important limitation. It requires the model to produce a sufficiently balanced distribution of correct and incorrect answers. When the correct-to-incorrect ratio deviates substantially from 50:50, balancing leaves few examples of each class, so the probe has little data with which to detect representational differences. This dependency means our findings are most reliable for models operating near their capability boundaries, where both outcomes occur with reasonable frequency. Olmo-3-Instruct on Sequential Growth (50 incorrect of 1,000) and Counting (10 incorrect of 1,000), and Olmo-3-Think on Conditional Probability (5 incorrect of 1,000), cannot be evaluated.

Other limits apply. Output length and truncation alone match the DeepSeek-Math-RL probe on Student Demographics (93\%) and Counting (79\% against 78\%), where it truncates 703 and 559 of 1,000 outputs at its 4,096-token context limit. Labels floor or round intermediate values. On Student Demographics, only 5 of 1,000 Olmo-3-Think answers are wrong if any integer within 1 of the exact answer counts as correct, so its probe mostly separates this labelling effect. On Sequential Growth, Olmo-3-Think splits between two readings of ``total amount'' (498 and 478 of 1,000), and its 100\% probe accuracy probably reflects which reading it used, although we did not test this directly.

\section{Layer-Wise Mean Ablations}

\label{sec:layer_ablation}
We aim to investigate the criticality each layer has upon the mathematical reasoning capabilities of DeepSeek-Math models through systematic activation patching. Our methodology employs mean ablation interventions to replace layer activations with their corresponding mean values computed from a reference dataset (GSM8K training data). This approach is motivated by the activation patching practices examined by Zhang and Nanda (2024).

\subsection{Experimental Setup}
We evaluate the \texttt{DeepSeek-Math-7B-Instruct} and \texttt{DeepSeek-Math-7B-RL} models on 20 GSM8K problems per model. For each layer $\ell \in \{0, 1, \ldots, L-1\}$, we replace the activation $h_{\ell}$ with its corresponding reference mean activation $\mu_{\ell}$ and measure the resulting change in accuracy.

\subsection{Evaluation Metric}
\textbf{Accuracy Drop (AD):} This metric quantifies the change in model accuracy relative to the baseline performance. For each layer $\ell$, we compute:
$
\mathrm{AD}_{\ell} = \mathrm{Acc}_{\text{base}} - \mathrm{Acc}_{\ell}^{\text{abl}},
$
where $\mathrm{Acc}_{\text{base}}$ is the baseline accuracy (without ablation), and $\mathrm{Acc}_{\ell}^{\text{abl}}$ is the accuracy measured when the activations at layer $\ell$ are replaced by $\mu_{\ell}$. Larger values of $\mathrm{AD}_{\ell}$ indicate higher importance of that layer in mathematical reasoning. Negative values mean accuracy rose under ablation.

\textbf{Pearson Correlation Coefficient (\textit{r}):} This statistic measures the linear correlation between layer depth and accuracy drop. A positive $r$ indicates that deeper layers are more critical to performance, while a value near zero implies no linear trend with depth.

\subsection{Implementation}
We extract final answers using pattern matching on the \verb|\boxed{...}| notation. All generations use fixed decoding parameters (temperature = 0.1, top\_p = 0.9). Prompts enforce structured, step-by-step reasoning to ensure that mathematical problem-solving processes are made explicit.

\subsection{Results and Analysis}

\begin{figure}[h]
    \centering
    \includegraphics[width=0.85\linewidth]{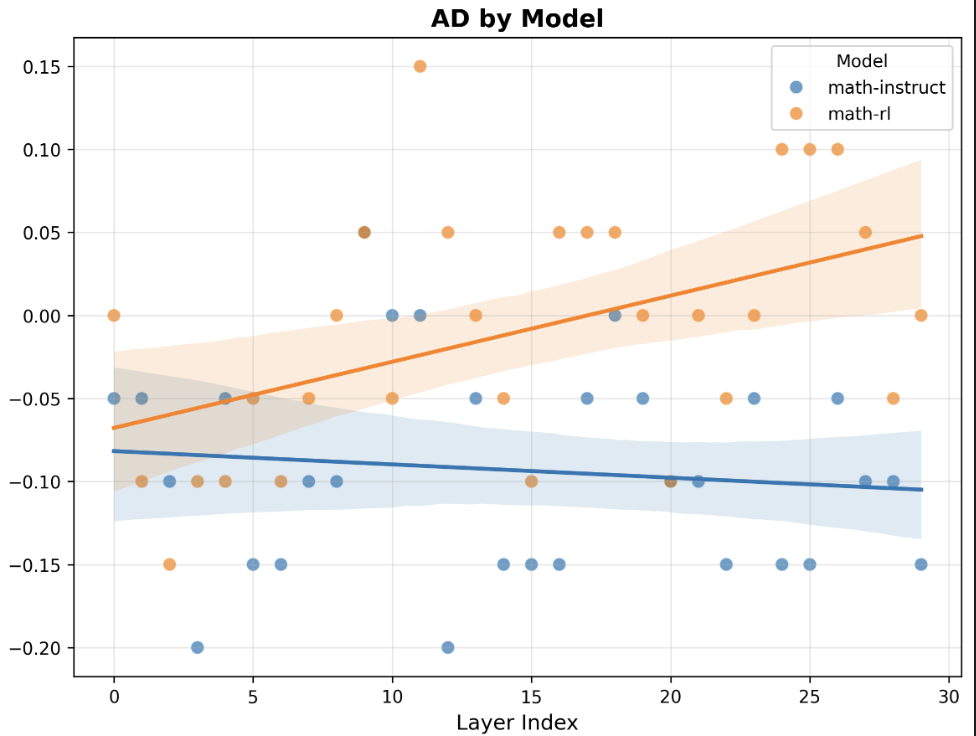}
    \caption{Accuracy Drop (AD) across layers for DeepSeek-Math-7B-Instruct and DeepSeek-Math-7B-RL.}
    \label{fig:ad_by_model}
\end{figure}

\subsection{Layer Criticality Patterns}
As shown in Figure~\ref{fig:ad_by_model}, the two DeepSeek variants exhibit different patterns. \texttt{DeepSeek-Math-7B-RL} (baseline acc. 70\%) exhibits a significant positive correlation between layer depth and intervention impact ($r = 0.47$, $p<0.01$), with AD ranging from $-0.15$ to $+0.15$. This suggests that deeper layers may become more critical for mathematical reasoning. In contrast, \texttt{DeepSeek-Math-7B-Instruct} (baseline accuracy: 65\%) demonstrates a weak negative correlation ($r = -0.11$, $p = 0.55$), with AD ranging from $-0.20$ to $+0.05$ and no clear trend across layers. AD is negative at 26 of 30 layers for \texttt{DeepSeek-Math-7B-Instruct} and 13 of 30 for \texttt{DeepSeek-Math-7B-RL}. With 20 problems per model, one problem is worth 0.05 of AD, so these estimates are coarse.

\subsection{Layer Criticality Interpretation}
These contrasting patterns may reflect different computational strategies shaped by training objectives.
The RL-trained model shows progressive deepening, which may indicate a hierarchical reasoning architecture.
Conversely, the SFT model shows no clear trend, and ablation raised its accuracy at most layers, so this experiment does not show how it distributes reasoning.

\subsection{Convergence and Divergence Points}
Both models exhibit similar AD across layers 0--10 (from $-0.20$ to $+0.05$), which may reflect shared mechanisms. Beyond layer 10, however, their trajectories diverge, which may reflect their different training methodology. This divergence may have implications for performance optimization and failure mode identification.

\section{Token-Count Variability in Mathematical Problem-Solving}
\label{sec:behavioural}

We investigate whether the probe differences between RL-trained reasoning language models and instruction-tuned models manifest in downstream behaviours such as token-count variability. Token-count variability is how much output token counts vary across the responses to one problem. We also measure whether token counts vary more on harder problems.

\subsection{Experimental Setup}

\textbf{Data and Methodology.} We evaluate the same four models on 300 problems from GSM8K-Platinum~\cite{vendrow2025largelanguagemodelbenchmarks}, generating 50 independent responses per problem per model (15,000 per model). We measure answer correctness, input tokens, and output tokens, including reasoning tokens for RL-trained reasoning language models. Full experimental details are in Appendix~\ref{app:token_variability_experimental_details}.

\textbf{Evaluation Metrics.} For each problem, we compute (1)~\textit{accuracy}, the proportion of responses producing correct answers, (2)~\textit{token coefficient of variation (CV)}, and (3)~\textit{median output tokens}. We compute CV across the 50 responses as $\mathrm{CV} = (\sigma_{\mathrm{tokens}} / \mu_{\mathrm{tokens}})(1 + 1/(4n))$, with the small-sample correction for $n$ responses, where $\sigma_{\mathrm{tokens}}$ and $\mu_{\mathrm{tokens}}$ are the standard deviation and mean of output token counts respectively. We use CV rather than raw standard deviation to enable fair comparison across model families with different baseline output lengths. Olmo-3-Think writes about 6$\times$ as many tokens as Olmo-3-Instruct, while DeepSeek-Math-RL writes about as many as DeepSeek-Math-Instruct. Medians are 162 for DeepSeek-Math-RL, 172 for DeepSeek-Math-Instruct, 1,834 for Olmo-3-Think, and 289 for Olmo-3-Instruct. Since all responses contain at least 37 tokens, the mean is never near zero, avoiding the instability that CV exhibits when $\mu \to 0$. We bin problems by accuracy. Figure bars show the mean CV over the problems in each bin, and error bars are 95\% $t$-intervals.

\subsection{Results: Variability Patterns}

\begin{figure*}[t]
\centering
\begin{subfigure}[b]{0.48\textwidth}
    \centering
    \includegraphics[width=\linewidth]{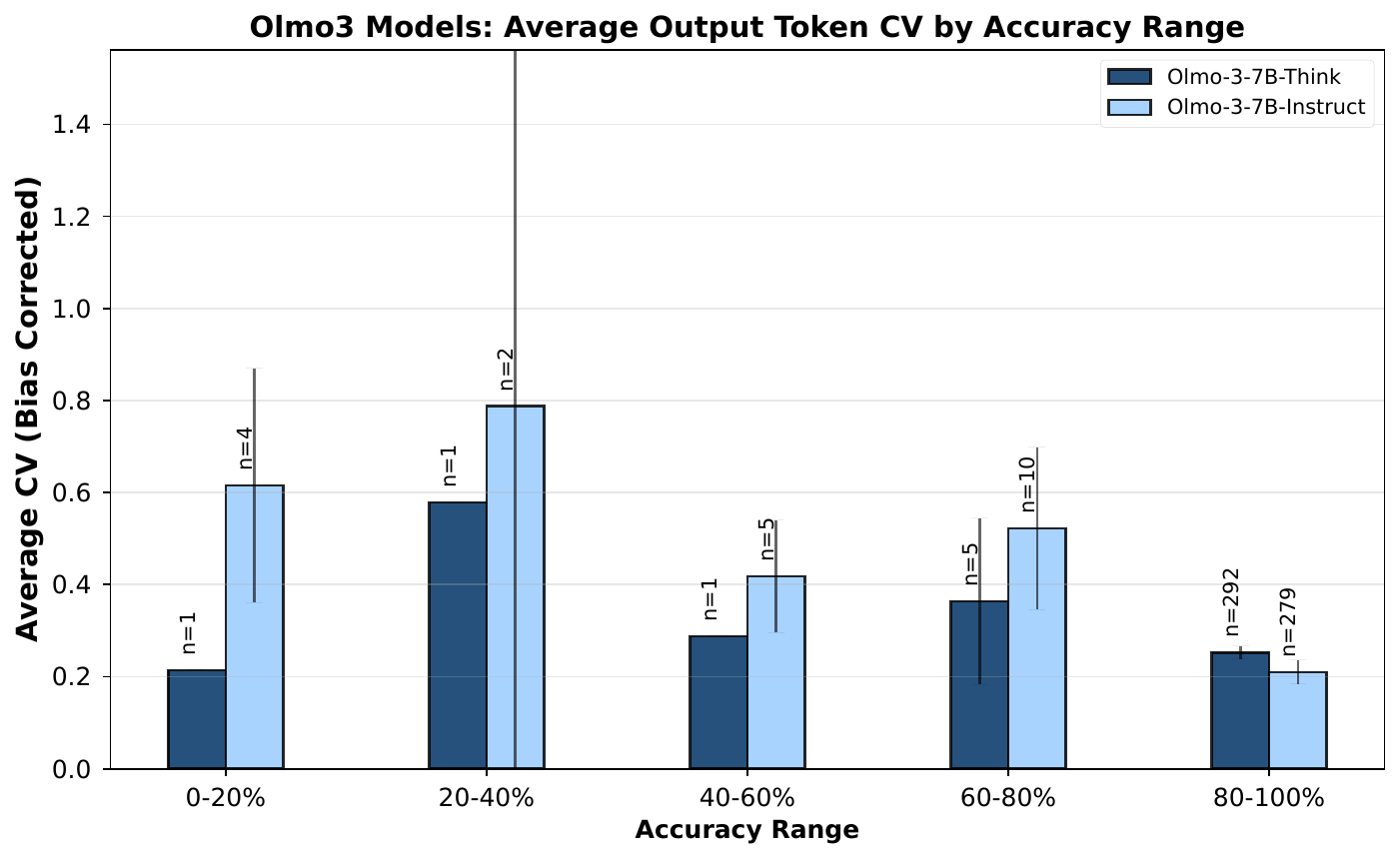}
    \caption{Olmo-3 models}
    \label{fig:qwen_cv_acc}
\end{subfigure}
\hfill
\begin{subfigure}[b]{0.48\textwidth}
    \centering
    \includegraphics[width=\linewidth]{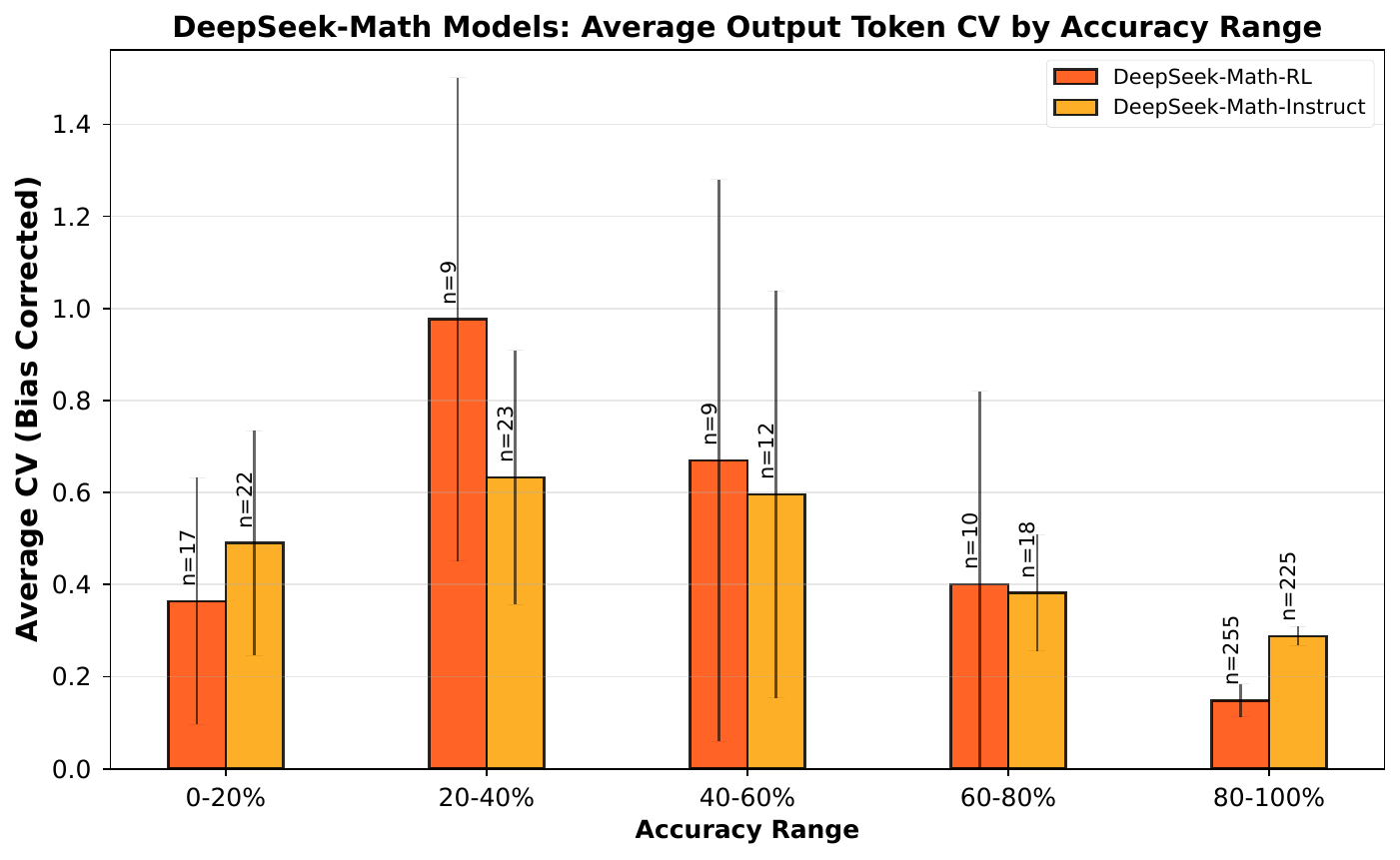}
    \caption{DeepSeek-Math models}
    \label{fig:deepseek_cv_acc}
\end{subfigure}
\caption{Token coefficient of variation by accuracy bin across model families. (a)~Olmo-3-Instruct has $\mathrm{CV} = 0.21$ at 80--100\% (279 problems) and $0.42$--$0.79$ in the other bins (2--10 problems each). Olmo-3-Think has $\mathrm{CV} = 0.25$ at 80--100\% (292 problems) and $0.21$--$0.58$ in the other bins (1--5 problems each). (b)~DeepSeek-Math variability peaks at 20--40\% and is lowest at 80--100\%. Bars include truncated responses (174 of 15,000 for DeepSeek-Math-RL, 26 for DeepSeek-Math-Instruct, 7 for Olmo-3-Think). Bins with few problems have wide intervals, bins with one problem have none, and $n$ labels give the problem count.}
\label{fig:combined_cv_acc}
\end{figure*}

\textbf{Variability peaks at 20--40\% accuracy in all four models.} Figure~\ref{fig:deepseek_cv_acc} reveals that DeepSeek-Math-RL peaks in the 20--40\% accuracy bin ($\mathrm{CV} \approx 0.98$), with $\mathrm{CV} \approx 0.36$ at 0--20\% and $\mathrm{CV} \approx 0.15$ at 80--100\%. Truncated responses drive most of the peak ($\mathrm{CV} \approx 0.41$ without them). DeepSeek-Math-Instruct shows higher variability than DeepSeek-Math-RL at 0--20\% ($\mathrm{CV} \approx 0.49$) and 80--100\% ($\mathrm{CV} \approx 0.29$). Figure~\ref{fig:qwen_cv_acc} shows Olmo-3-Instruct at $\mathrm{CV} = 0.21$ and Olmo-3-Think at $\mathrm{CV} = 0.25$ in the 80--100\% bin. DeepSeek-Math-RL, DeepSeek-Math-Instruct, and Olmo-3-Instruct are lowest at 80--100\%, and Olmo-3-Think is lowest on its one problem at 0--20\%.

\textbf{Partial evidence for compression theory.} Our findings partially support \citet{lee2025llmscompresschainofthoughttoken}'s token complexity framework. DeepSeek-Math-RL exhibits high variability at 20--40\% accuracy, meaning one problem elicits vastly different response lengths, although truncated responses drive most of this. Even at high accuracy, where models should reliably compress, non-trivial variability persists in both DeepSeek-Math models, indicating systematic deviation from optimal compression. The Olmo-3 family does not show consistent low variability. Olmo-3-Instruct ranges from $\mathrm{CV} = 0.21$ to $0.79$ across bins, and Olmo-3-Think from $0.21$ to $0.58$. In all four models, CV tends to rise as accuracy falls (Spearman $\rho$ from $-0.13$ to $-0.45$), although Olmo-3-Think has only 8 problems at or below 80\%, and for DeepSeek-Math-Instruct the trend depends on truncated responses ($\rho = -0.09$, $p = 0.10$ without them).

\textbf{RL training and adaptive allocation.} DeepSeek-Math-RL exhibits lower variability than DeepSeek-Math-Instruct in the 80--100\% bin ($\mathrm{CV} \approx 0.15$ against $0.29$), which holds 255 and 225 of the 300 problems respectively. The other bins hold few problems and have overlapping intervals. In contrast, Olmo-3-Think varies more than Olmo-3-Instruct on the 276 problems in the 80--100\% bin for both models ($\mathrm{CV} = 0.24$ against $0.20$, paired $t$-test $p = 0.0015$, chosen after the first 50 problems). Over all 300 problems the means do not differ significantly ($0.26$ against $0.23$, $p = 0.095$), because Olmo-3-Instruct varies more on the other 24 ($0.57$ against $0.38$). Olmo-3-Think is still higher on 207 of the 300 (Wilcoxon $p < 0.001$). Only the DeepSeek pair contrasts RL with SFT, so these results rest on one pair.

\section{Conclusion}
We investigated the mechanistic basis of reinforcement learning's success in mathematical reasoning through integrated behavioural-mechanistic analysis.
Our findings give a mixed picture.
Linear probes tend to predict answer correctness more accurately in RL-trained reasoning language models than in instruction-tuned models, but confounds limit what this shows.
Specifically, output length alone matches the DeepSeek-Math-RL probe on two templates, and DeepSeek-Math-RL is level with its SFT counterpart in probe accuracy on one other and lower on another, while Olmo-3-Think scores higher than Olmo-3-Instruct on one of the two templates where both can be evaluated.
Mean ablation on the DeepSeek-Math pair suggests that deeper layers may matter more in DeepSeek-Math-RL, but it does not test these representations.
Token-count variability analysis reveals model-dependent patterns.
DeepSeek-Math-RL exhibits lower variability than its SFT counterpart in the 80--100\% accuracy bin, which holds most problems, while Olmo-3-Think varies more than Olmo-3-Instruct in that bin.
Only the DeepSeek pair contrasts RL with SFT, so this variability result rests on one pair.

Multiple promising avenues emerge from this work.
Designing reward structures that explicitly incentivise adaptive token allocation could better exploit RL's potential.
Our layer-wise analysis focused on answer correctness.
Extending probing to intermediate reasoning steps could reveal how multi-step solutions are constructed and validated.
Scaling this analysis to larger models and diverse reasoning domains (code generation, scientific reasoning) would test whether our findings generalise beyond mathematical problem-solving.
More broadly, developing methods to detect and measure representation quality during training could enable real-time assessment of model reliability, a critical need for deploying reasoning models in high-stakes domains.

\section*{Correction Note}

This version corrects arXiv:2607.26119v1. An audit of v1 against its code found errors in the probing and token-count variability results, so we reran both experiments, and the results here replace those in v1. The layer-wise ablation experiment has not been rerun. Its figure is unchanged. This version corrects the text's reading of it: the AD range for layers 0--10, the claim of higher early-layer impact, and the layer where the trajectories diverge (10 rather than 15). It also adds the number of layers with negative AD, where accuracy rose, and softens the claims drawn from this figure.

\textbf{Linear probes.}
\begin{itemize}
    \item v1 described 50/50 balancing, but its code used all 1,000 problems per template at each model's natural class ratio. The rerun balances classes for each model and template.
    \item v1 extracted activations from the completion alone. The rerun includes the prompt in the forward pass.
    \item v1 said all answers follow \texttt{\textbackslash boxed\{\}}. DeepSeek-Math-RL boxes 2.0\% of its outputs, so the rerun probes it at the final ``The answer is:'' when no box appears.
    \item v1's accuracy ranges (83--98\% and 75--90\%) and emergence layers did not match its stored results. v1's Figure~1 showed one template without saying so.
    \item v1's Sequential Growth figure was a copy of its Cost Calculation figure. All four probe figures are regenerated.
    \item This version adds the output-length baseline, the label and question-reading limits, and model accuracy for each template (Table~\ref{tab:model_accuracy}).
\end{itemize}

\textbf{Token-count variability.}
\begin{itemize}
    \item v1 stated 15,000 responses per model. Its figure used 50 problems and 2,500 responses, or 1,500 for Olmo-3-Instruct. The rerun uses 300 problems and 15,000 responses per model.
    \item v1 described 50 independent responses per problem, but it set no seed, so whole runs repeated. Only 22 of 50 runs were distinct for DeepSeek-Math-RL, 18 of 50 for DeepSeek-Math-Instruct, 7 of 50 for Olmo-3-Think, and 7 of 30 for Olmo-3-Instruct. This biased v1's CV low. The rerun seeds each problem, which gives a mean of 46 to 50 distinct responses per problem.
    \item In v1, DeepSeek-Math-RL received the bare question with no instruction, DeepSeek-Math-Instruct ran at $T = 0.7$ rather than the stated $0.6$, and the Olmo-3 models ran at the default top-p of $1.0$ rather than the stated $0.95$. Its grader marked answers wrong when the reference used a thousands separator. The rerun uses one prompt format, $T = 0.6$, top-p $0.95$, and numeric grading for all models.
    \item v1 reported substantially higher variability for DeepSeek-Math-RL than for DeepSeek-Math-Instruct in all bins. In the rerun it is lower in the 80--100\% bin ($\mathrm{CV} \approx 0.15$ against $0.29$).
    \item v1 reported $\mathrm{CV} = 0.1$--$0.125$ for the Olmo-3 models. In the rerun, Olmo-3-Instruct ranges from $0.21$ to $0.79$ across bins, and Olmo-3-Think from $0.21$ to $0.58$.
\end{itemize}

\textbf{Model descriptions.} v1 described DeepSeek-Math-RL as trained from verifiable rewards and Olmo-3-Instruct as an SFT model. DeepSeek-Math-RL uses GRPO with a learned reward model, and Olmo-3-Instruct also includes DPO and RLVR, so only the DeepSeek pair contrasts RL with SFT.

\textbf{Conclusions.} The probe advantage of RL-trained reasoning language models is less consistent across templates and partly confounded, and DeepSeek-Math-RL no longer shows higher token-count variability than DeepSeek-Math-Instruct. This version no longer claims that RL-trained models develop representations in earlier layers, that ablation confirms these representations are functionally critical, that reasoning-specific objectives are key to clearer representations, or that Olmo-3 models generate consistently across difficulty levels. Probe accuracy also does not track model accuracy (Spearman $\rho = 0.02$), contrary to v1's hypothesis.

\appendix
\section{Additional Token-Count Variability Experimental Details}
\label{app:token_variability_experimental_details}

\textbf{Dataset:} 300 problems randomly sampled from the GSM8K-Platinum test split~\cite{vendrow2025largelanguagemodelbenchmarks} (seed=42).

\textbf{Generation parameters:} All models use temperature $T = 0.6$ and top-p $p = 0.95$. Max tokens is 3800 for DeepSeek-Math-Instruct and DeepSeek-Math-RL, and 32768 for Olmo-3-Instruct and Olmo-3-Think.

\textbf{Model inference:} We use vLLM~\cite{kwon2023efficient} on a single NVIDIA H100 GPU for efficient execution. The last 250 Olmo-3-Think problems instead ran on a single NVIDIA RTX PRO 6000 with vLLM's async scheduling.

\textbf{Prompt template:}
\begin{verbatim}
{question}
Please reason step by step, and
put your final answer within \boxed{}.
\end{verbatim}
Each model's chat template wraps this prompt. The DeepSeek-Math template adds ``User:'' and ``Assistant:''. The Olmo-3 templates add their default system text, and the Olmo-3-Think template also opens a think block.

\section{Additional Details for Linear Probing}
\label{sec:linprobeexperiment}

\textbf{Problem templates:} Each template instantiates a word problem with randomised numerical parameters and a fixed logical structure:
\begin{enumerate}
    \item \textit{Conditional probability}: Calculate the probability of turning in homework given sequential conditional events (substitute teacher, class extension, personal extension). Requires probability multiplication and complementary probability computation. Answers range from 8 to 50\%. Sample question: Yasmine is trying to decide whether they really need to do their homework. There's a 70\% chance that tomorrow they'll have a substitute teacher who won't collect the homework. Even if the normal teacher comes in, there's a 60\% chance she'll give everyone an extension. Even if the whole class doesn't get an extension, there's a 25\% chance Yasmine can convince the teacher their dog ate their assignment and get a personal extension. What is the percentage chance that Yasmine will actually have to turn in their homework tomorrow? \textit{(Answer: 9\%)}
    
    \item \textit{Student demographics}: Given total students, age threshold, and gender ratios stratified by age group, compute total female students. Requires division, fraction multiplication, and subtraction. Answers range from 220 to 3,986 students. Sample question: Brook Hills High School currently enrolls 4,374 students. Half of these students are over 18 years old, and one-fifth of the students over 18 years old are male. The remaining half of the students are under 18 years old, and 2/5 of the students under 18 are male. In total, how many female students are enrolled at this school? \textit{(Answer: 3,063 students)}
    
    \item \textit{Sequential growth}: Given initial water flow and multiplicative and additive growth rules over days, compute the final quantity. Requires tracking state across time steps with doubling and addition. Answers range from 7,057 to 25,513 gallons. Sample question: The amount of water passing through a river at one point in time is 5,904 gallons. After a day of heavy rain, the amount of water passing through the river doubles at the same point. If the volume of water passing through the river at that point increases by 7,202 gallons on the third day, calculate the total amount of water passing through the river at that point. \textit{(Answer: 19,010 gallons)}
    
    \item \textit{Counting with unit conversion}: Track brownies (in dozens) received and consumed across multiple events, then convert to individual items. Requires dozen-to-unit conversion, fraction addition and subtraction, and summation. Answers range from 1 to 231 brownies. Sample question: Quentin wanted brownies for her birthday. She made a batch for herself; nine dozen Nut Brownies brownies. At her office, they threw her a party and sent her home with 9/10 dozen brownies. When she arrived home, her friends were there to throw her a surprise party and had 4 dozen brownies waiting. During the party, 2 2/10 dozen brownies were eaten. How many individual brownies did Quentin have left over from the entire day? \textit{(Answer: 140 brownies)}
    
    \item \textit{Cost calculation}: Given a base price and dependent pricing rules (e.g., ``leather seats cost one-third of the king cab upgrade''), compute the total cost. Requires chained fraction operations and summation. Answers range from $\$34,490$ to $\$66,846$. Sample question: Bill is ordering a new truck. He has decided to purchase a two-ton truck with several added features: a king cab upgrade, a towing package, leather seats, running boards, and the upgraded exterior light package. The base price of the truck is \$42,572, and the other features are at extra cost. The king cab is an extra \$6,890, leather seats are one-third the cost of the king cab upgrade, running boards are \$500 less than the leather seats, and the upgraded exterior light package is \$1,724. What is the total cost of Bill's new truck, in dollars? \textit{(Answer: \$55,278)}
\end{enumerate}

Answers range from 1 to 66,846 across templates. A sixth template (page folding) was also run, but we do not report it because its 1,000 rows hold only 15 distinct questions, so train and test share questions. Conditional Probability has 871 distinct questions, so some of its questions also appear in both train and test.

Each question includes the instruction ``Please reason step by step, and put your final answer within \texttt{\textbackslash boxed\{answer\}} as an integer.'' The system prompt is ``Solve the following problem'', and the user turn is the question followed by a second instruction, ``Please reason step by step, and put your final answer within \texttt{\textbackslash boxed\{\}}.'' All models are therefore prompted to reason step by step. Each model's chat template wraps this prompt, and the Olmo-3-Think template appends a functions notice to the system text and opens a think block.

\textbf{Reproducibility:} Code and final results are provided in the GitHub link.

\subsection*{Per-Template Probe Accuracy Results}

The following figures show test set probe accuracy on balanced data across all transformer layers for four of the five synthetic problem templates. Each figure displays the models that can be evaluated on that template.

\begin{figure}[htbp]
    \centering
    \includegraphics[width=0.9\linewidth]{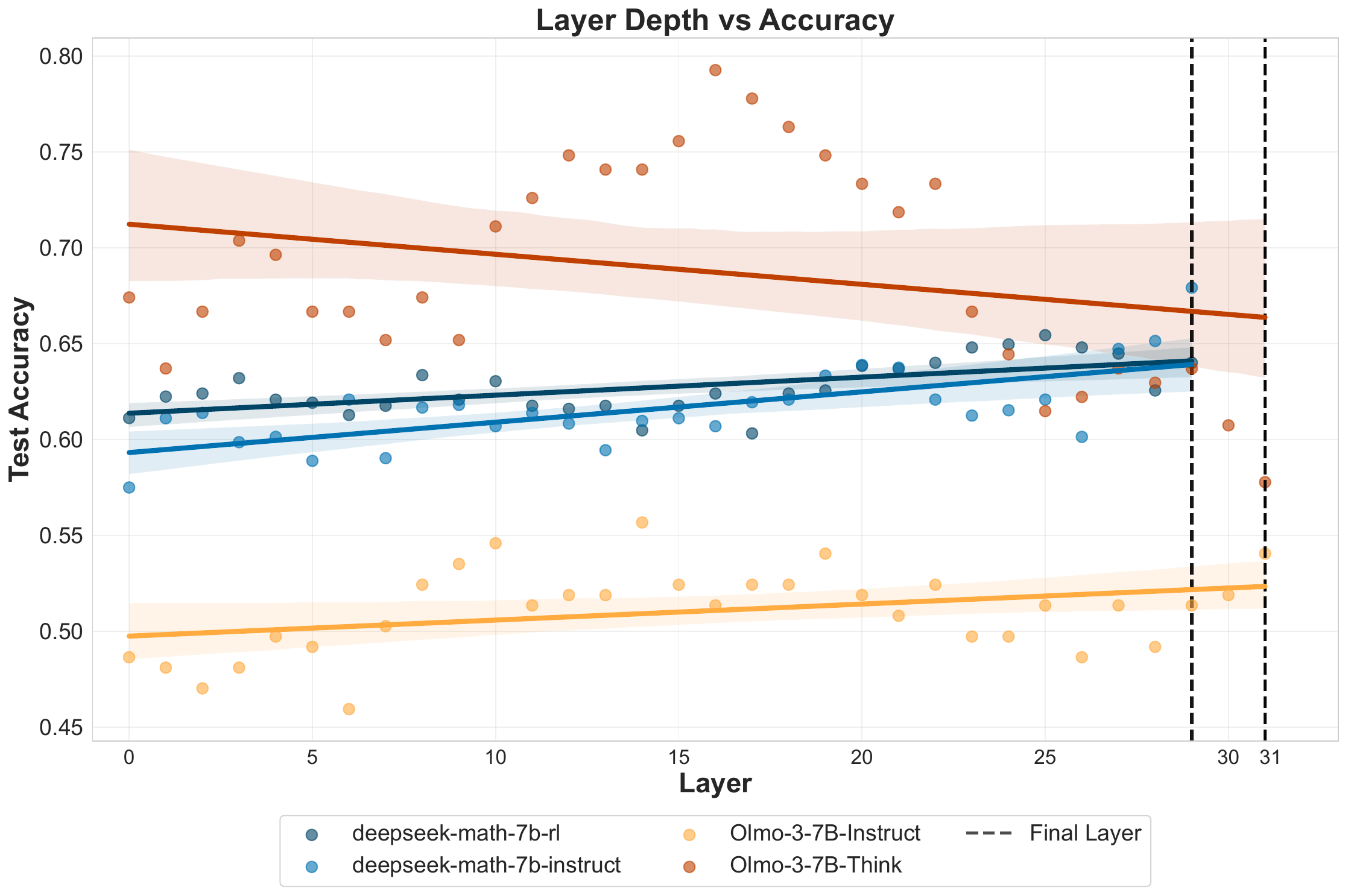}
    \caption{
    Probing classification accuracy vs. transformer layer.
    Cost Calculation template: DeepSeek-Math-RL, DeepSeek-Math-Instruct, Olmo-3-Think, Olmo-3-Instruct.
    }
    \label{fig:cost_calc}
\end{figure}

\begin{figure}[htbp]
    \centering
    \includegraphics[width=0.9\linewidth]{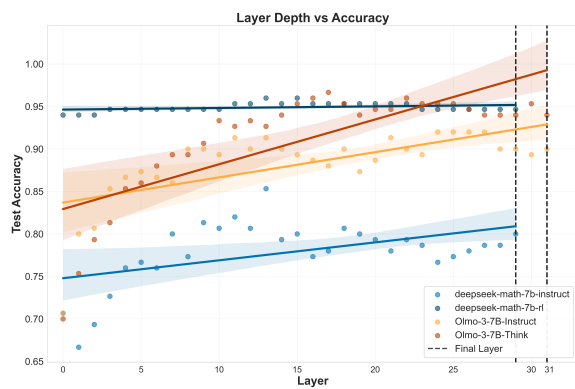}
    \caption{
    Probing classification accuracy vs. transformer layer.
    Student Demographics template: DeepSeek-Math-RL, DeepSeek-Math-Instruct, Olmo-3-Think, Olmo-3-Instruct.
    }
    \label{fig:student_pop}
\end{figure}

\begin{figure}[htbp]
    \centering
    \includegraphics[width=0.9\linewidth]{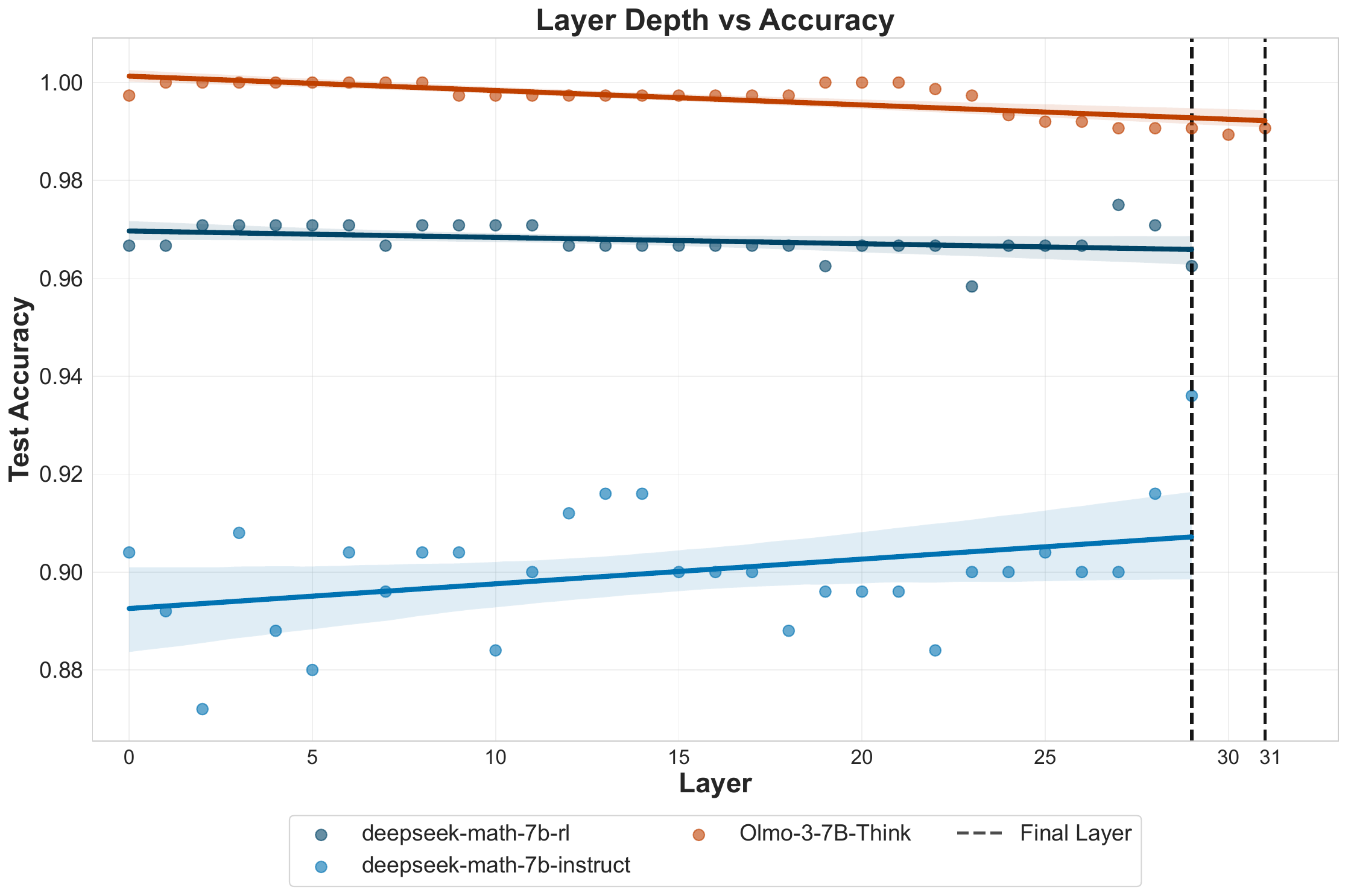}
    \caption{
    Probing classification accuracy vs. transformer layer.
    Sequential Growth template: DeepSeek-Math-RL, DeepSeek-Math-Instruct, Olmo-3-Think. Olmo-3-Instruct has too few incorrect answers to evaluate (50 of 1,000).
    }
    \label{fig:seq_grow}
\end{figure}

\begin{figure}[htbp]
    \centering
    \includegraphics[width=0.9\linewidth]{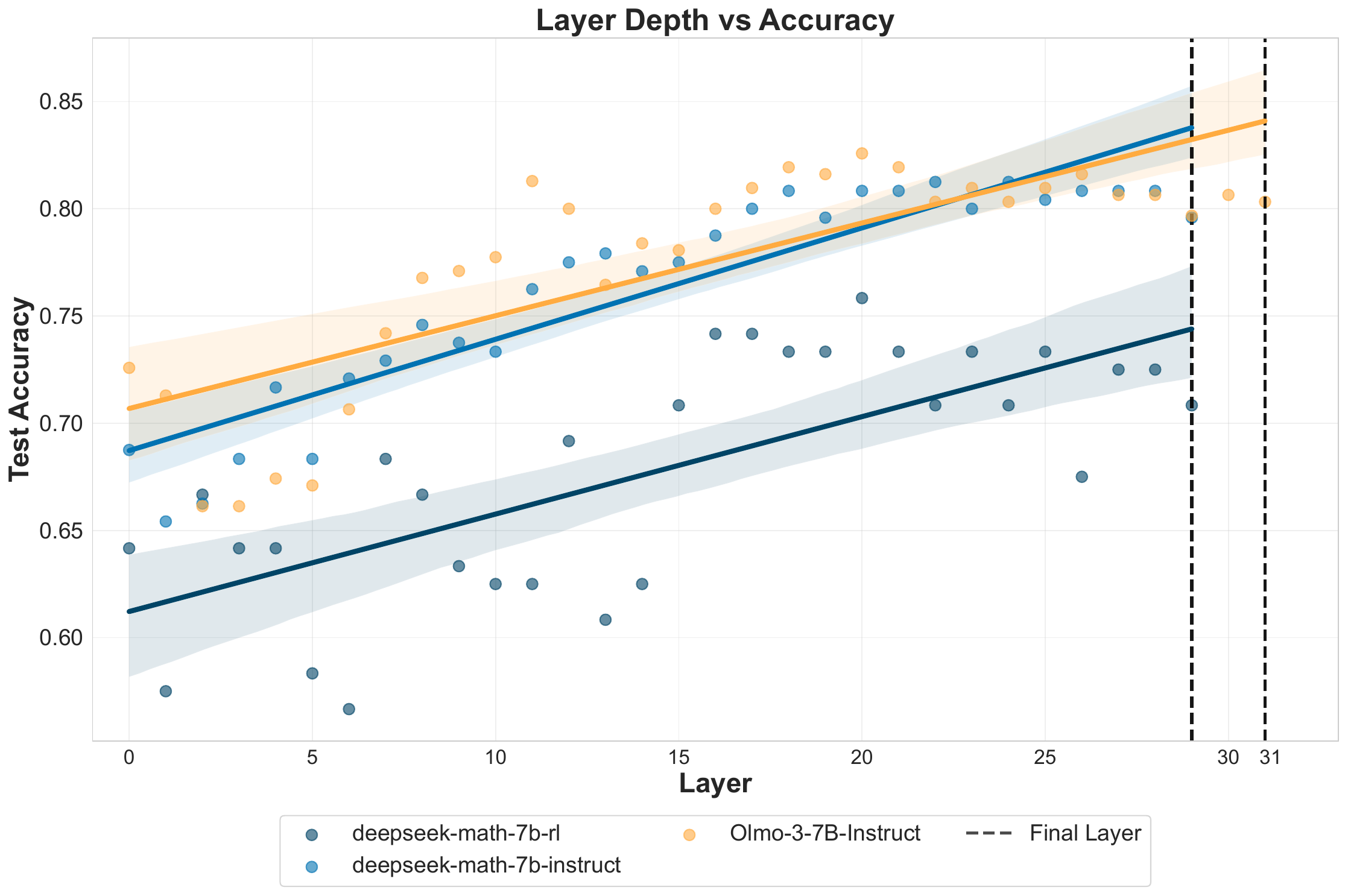}
    \caption{
    Probing classification accuracy vs. transformer layer.
    Conditional Probability template: DeepSeek-Math-RL, DeepSeek-Math-Instruct, Olmo-3-Instruct. Olmo-3-Think has too few incorrect answers to evaluate (5 of 1,000).
    }
    \label{fig:cond_prob}
\end{figure}

\bibliography{aaai2026}

\end{document}